\documentclass[11pt,letterpaper]{article}
\usepackage{emnlp2017}
\usepackage{times}
\usepackage{latexsym}

\usepackage{algorithm}
\usepackage{algorithmic}

\usepackage{amsmath}
\usepackage{amssymb}
\DeclareMathOperator*{\argmax}{arg\,max}
\usepackage{graphicx}
\usepackage{subcaption}
\usepackage{multirow}
\usepackage{siunitx}
\usepackage{booktabs}
\usepackage{soul}
\usepackage{paralist}
\usepackage{xspace}

\newcommand{\pal}{\textsc{Pal}\xspace}
\newcommand{\palb}{\textsc{Pal}$_b$\xspace}
\newcommand{\palm}{\textsc{Pal}$_m$\xspace}
\newcommand{\palc}{\textsc{Pal}$_c$\xspace}

\emnlpfinalcopy

\def\emnlppaperid{679}

\usepackage{url}

\usepackage{xcolor}

\usepackage[normalem]{ulem}

\makeatletter
\newcounter{ALC@tempcntr}
\makeatother

\title{Learning how to Active Learn: \\ A Deep Reinforcement Learning Approach}

\author{Meng Fang \and Yuan Li \and Trevor Cohn \\
School of Computing and Information Systems\\
The University of Melbourne \\ \texttt{meng.fang@unimelb.edu.au,~yuanl4@student.unimelb.edu.au,} \\ \texttt{t.cohn@unimelb.edu.au}}

\date{}

\begin{document}

\maketitle

\begin{abstract}
Active learning aims to select a small subset of data for annotation such that a classifier learned on the data is highly accurate. This is usually done using heuristic selection methods, however the effectiveness of such methods is limited and moreover, the performance of heuristics varies between datasets.
To address these shortcomings, we introduce a novel formulation by reframing the active learning as a reinforcement learning problem and explicitly learning a data selection policy, where the policy takes the role of the active learning heuristic. 
Importantly, our method allows the selection policy learned using simulation on one language to be transferred to other languages.
We demonstrate our method using cross-lingual named entity recognition, observing uniform improvements over traditional active learning.
\end{abstract}

\section{Introduction}
For most Natural Language Processing (NLP) tasks, obtaining sufficient annotated text for training accurate models is a critical bottleneck. 
Thus active learning has been applied to NLP tasks to minimise the expense of annotating data~\cite{thompson1999active,tong2001support,settles2008analysis}.
Active learning aims to reduce cost by identifying a subset of unlabelled data for annotation, which is selected to maximise the accuracy of a supervised model trained on the data~\cite{settles2010active}.
There have been many successful applications to NLP, e.g., \newcite{tomanek2007approach} used an active learning algorithm for CoNLL corpus to get an F$_1$ score $84\%$ with a reduction of annotation cost of about $48\%$.
In prior work most active learning algorithms are designed for English based on heuristics, such as using uncertainty or informativeness.
There has been comparatively little work done about how to learn the active learning strategy itself.

It is no doubt that active learning is extremely important for other languages, particularly low-resource languages, where annotation is typically difficult to obtain, and annotation budgets more modest~\cite{garrette2013learning}.
Such settings are a natural application for active learning, however there is little work to this end.
A potential reason is that most active learning algorithms require a substantial `seed set' of data for learning a basic classifier, which can then be used for active data selection.
However, given the dearth of data in the low-resource setting, this assumption can make standard approaches infeasible.

In this paper,\footnote{Source code available at \url{https://github.com/mengf1/PAL}} we propose \pal, short for \emph{Policy based Active Learning},
a novel approach for learning a dynamic active learning strategy from data.
This allows for the strategy to be applied in other data settings, such as cross-lingual applications. Our algorithm does not use a fixed heuristic, but instead learns how to actively select data, formalised as a reinforcement learning (RL) problem. An intelligent agent must decide whether or not to select data for annotation in a streaming setting, where the decision policy is learned using a deep Q-network~\cite{mnih2015human}. The policy is informed by observations including sentences' content information, the supervised model's classifications and its confidence. Accordingly, a rich and dynamic policy can be learned for annotating new data based on the past sequence of annotation decisions.


Furthermore, in order to reduce the dependence on the data in the target language, which may be low resource, we first learn the policy of active learning on another language and then transfer it to the target language. It is easy to learn a policy on a high resource language, where there is plentiful data, such as English. We use cross-lingual word embeddings to learn compatible data representations for both languages, such that the learned policy can be easily ported into the other language.

Our work is different for prior work in active learning for NLP. Most previous active learning algorithms developed for NER tasks is based on one language and then applied to the language itself. Another main difference is that many active learning algorithms use a fixed data selection heuristic, such as uncertainty sampling~\cite{settles2008analysis,stratos2015simple,zhang2016name}. However, in our algorithm, we implicitly use uncertainty information as one kind of observations to the RL agent.


The remainder of this paper is organised as follows. In Section 2, we briefly review some related work. In Section 3, we present active learning algorithms, which cross multiple languages. The experimental results are presented in Section 4. We conclude our work in Section 5.

\section{Related work}
As supervised learning methods often require a lot of training data, active learning is a technique that selects a subset of data to annotate for training the best classifier.
Existing active learning (AL) algorithms can be generally considered as three categories: 1) uncertainty sampling~\cite{lewis1994sequential,tong2001support}, which selects the data about which the current classifier is the most uncertain; 2) query by committee~\cite{seung1992query}, which selects the data about which the ``committee'' disagree most; and 3) expected error reduction~\cite{roy2001toward}, which selects the data that can contribute the largest model loss reduction for the current classifier once labelled. Applications of active learning to NLP include text classification~\cite{mccallumzy1998employing,tong2001support}, relation classification~\cite{Qian2014BilingualAL}, and structured prediction~\cite{shen2004multi,settles2008analysis,stratos2015simple,fang2017model}.
\citeauthor{Qian2014BilingualAL} used uncertainty sampling to jointly perform on English and Chinese. ~\citeauthor{stratos2015simple} and \citeauthor{zhang2016name} deployed uncertainty-based AL algorithms for languages with the minimal supervision.

Deep reinforcement learning (DRL) is a general-purpose framework for decision making based on representation learning. Recently, there are some notable examples include deep Q-learning~\cite{mnih2015human}, deep visuomotor policies~\cite{levine2016end}, attention with recurrent networks~\cite{ba2014multiple}, and model predictive control with embeddings~\cite{watter2015embed}. Other important works include massively parallel frameworks~\cite{nair2015massively}, dueling architecture~\cite{wang2015dueling} and expert move prediction in the game of Go~\cite{maddison2014move}, which produced policies matching those of the Monte Carlo tree search programs, and squarely beaten a professional player when combined with search~\cite{silver2016mastering}. 
DRL has been also studied in NLP tasks. For example, recently, DRL has been studied for information extraction problem~\cite{narasimhan2016improving}. They designed a framework that can decide to acquire external evidence and the framework is under the reinforcement learning method. 
However, there has been fairly little work on using DRL to learn active learning strategies for language processing tasks, especially in cross-lingual settings.

Recent deep learning work has also looked at transfer learning~\cite{bengio2012deep}. 
More recent work in deep learning has also considered transferring policies by reusing policy parameters
between environments~\cite{parisotto2015actor,rusu2016progressive},
using either regularization or novel neural network architectures, though this work has not looked at transfer active learning strategies between languages with shared feature space in state.




\section{Methodology}

We now show how active learning can be formalised as as a decision process, and then show how this allows for the active learning selection policy to be learned from data using deep reinforcement learning. Later we introduce a method for transferring the policy between languages.

\subsection{Active learning as a decision process}

Active learning is a simple technique for labelling data, which involves first selecting some instances from an unlabelled dataset, which are then annotated by a human oracle, which is then repeated many times until a termination criterion is satisfied, e.g., the annotation budget is exhausted.
Most often the selection function is based on the predictions of a trained model, which has been fit to the labelled dataset at each stage in the algorithm, where datapoints are selected based on the model's predictive uncertainty \cite{lewis1994sequential}, or divergence in predictions over an ensemble \cite{seung1992query}.
The key idea of these methods is to find the instances on which the model is most likely to make errors, such that after their labelling and inclusion in the training set, the model becomes more robust to these types of errors on unseen data.

The steps in active learning can be viewed as a decision process, a means of formalising the active learning algorithm as a sequence of decisions, where the stages of active learning correspond to the state of the system.
Accordingly, the \emph{state} corresponds to the selected data for labelling and their labels, and each step in the active learning algorithm corresponds to a selection \emph{action}, wherein the heuristic selects the next items from a pool.
This process terminates when the budget is exhausted.

Effectively the active learning heuristic is operating as a decision \emph{policy}, a form of function taking as input the current state --- comprising the labelled data, from which a model is trained --- and a candidate unlabelled data point --- e.g., the model uncertainty.  This raises the opportunity to consider general policy functions, based on the state and data point inputs, and resulting in a labelling decision, and, accordingly a mechanism for learning such functions from data. We now elaborate on the components of this process, namely the formulation of the decision process, architecture of the policy function,  and means of  learning the decision policy automatically from data.



\begin{figure}[t]
\hspace{-2ex}
	\includegraphics[width=0.55\textwidth]{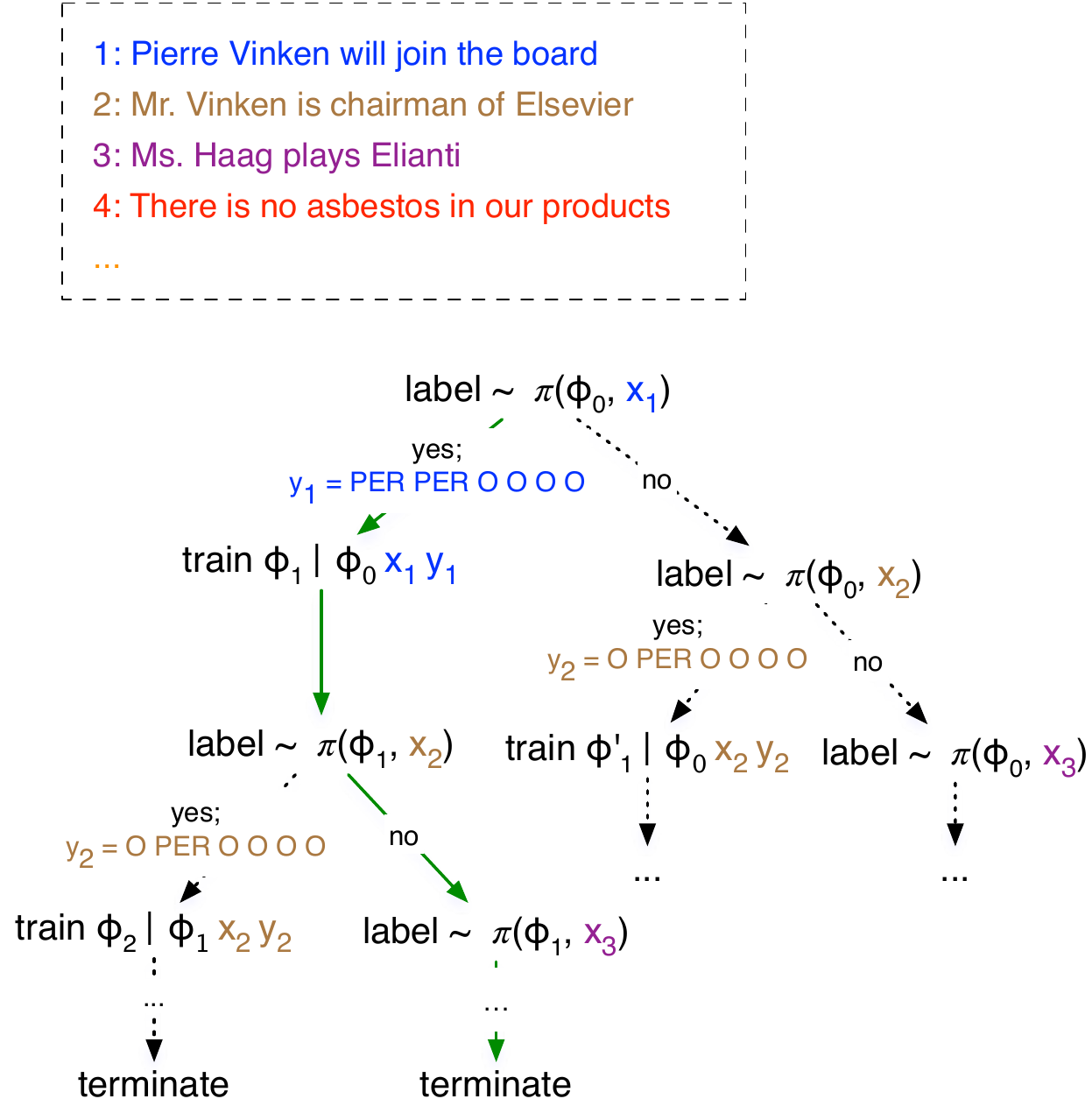}
	\caption{Example illustrating sequential active learning as a Markov Decision process. Data arrives sequentially, and at each time the active learning policy, $\pi$, must decide whether it should be labelled or not, based on the state which includes a predictive model parameterised by $\phi$, and an unlabelled data instance $\mathbf{x}$. The process continues until termination, e.g., when the annotation budget is exhausted. The solid green path shows the maximum scoring decision sequence.}
	\label{fig-mdp}
\end{figure}

\subsection{Stream-based learning}

 


For simplicity, we make a streaming assumption, whereby unlabelled data (sentences) arrive in a stream~\cite{lewis1994sequential}.\footnote{This is different to pool-based active learning, where one of several options is chosen for annotation. Our setup permits simpler learning, while remaining sufficiently general.}  As each instance arrives, an agent must decide the action to take, namely whether or not the instance should be manually annotated.
This process is illustrated in Figure~\ref{fig-mdp}, which illustrates the space of decision sequences for a small corpus.
As part of this process, a separate model, $p_\phi$, is trained on the labelled data, and updated accordingly as the labelled dataset is expanded as new annotations arrive. 
This model is central to the policy for choosing the labelling actions at each stage, and for determining the reward for a sequence of actions.

This is a form of Markov Decision Process (MDP), which allows the learning of a policy that can dynamically select instances that are most informative. 
As illustrated in Figure~\ref{fig-mdp} at each time, the agent observes the current state $s_i$ which includes the sentence $\mathbf{x}_i$, and the learned model $\phi$. The agent selects a binary action $a_i$, denoting whether to label $\mathbf{x}_i$,  according to the policy $\pi$. For $a_i=1$, the corresponding sentence is labelled and added to the labelled data, and the model $p_\phi$ updated to include this new training point. The process then repeats, terminating when either the dataset is exhausted or a fixed annotation budget is reached. 
After termination a reward  is computed based on the accuracy of the final model,~$\phi$. 
We represent the MDP framework as a tuple $\left< S, A, Pr(s_{i+1}|s_{i},a), R \right>$, where $S=\{s\}$ is the space of all possible states, $A=\{0,1\}$ is the set of actions, $R(s,a)$ is the reward function, and $Pr(s_{i+1}|s_{i},a)$ is the transition function.

\subsubsection{State}\label{sec-state}
The state at time $i$ comprises the candidate instance being considered for annotation and the labelled dataset constructed in steps $1 \ldots i$. We represent the state using a continuous vector, using the concatenation of the vector representation of $\mathbf{x}_i$, and outputs of the model $p_\phi$ trained over the labelled data. These outputs use both the predictive marginal distributions of the model on the instance, and a representation of the model's confidence. We now elaborate on each component.

\paragraph{Content representation} A key input to the agent is the content of the sentence, $\mathbf{x}_i$, which we encode using a convolutional neural network to arrive at a fixed sized vector representation, following~\newcite{kim2014convolutional}. This involves embedding each of the $n$ words in the sentence to produce a matrix $X_i = \{x_{i,1},x_{i,2},\cdots,x_{i,n}\}$, after which a series of wide convolutional filters is applied, 
using multiple filters with different gram sizes.
Each filter uses a linear transformation with a rectified linear unit activation function.
Finally the filter outputs are merged using a max-pooling operation to yield a hidden state $\mathbf{h}_c$, which is used to represent the sentence.

\paragraph{Representation of marginals}
The prediction outputs of the training model, $p_{\phi}(\mathbf{y} | \mathbf{x}_i)$, are central to all active learning heuristics, and accordingly, we include this in our approach. In order to generalise existing techniques, we elect to use the predictive marginals directly, rather than only using statistics thereof, e.g., entropy.
This generality  allows for different and more nuanced concepts to be learned, including patterns of probabilities that span several adjacent positions in the sentence (e.g., the uncertainty about the boundary of a named entity).


\begin{figure}[t]
	\centering
	\includegraphics[width=0.5\textwidth]{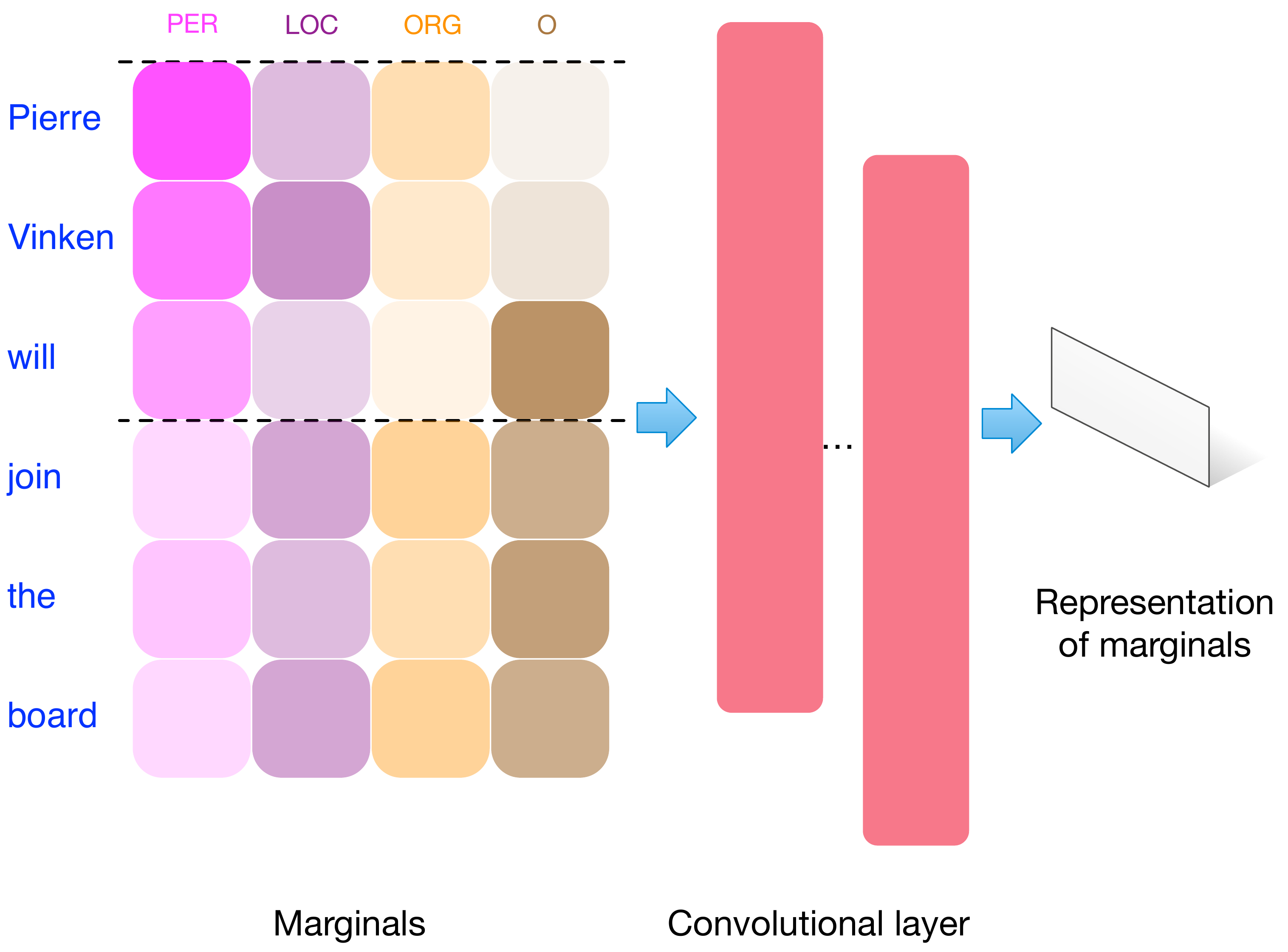}
	\caption{The architecture for representing predictive marginal distributions, $p_{\phi}(\mathbf{y} | \mathbf{x}_i)$, as a fixed dimensional vector, to form part of the MDP state.}
	\label{fig-cnn}
\end{figure}

We use another convolutional neural network to process the predictive marginals, as shown in Figure~\ref{fig-cnn}. The convolutional layer contains $j$ filters with ReLU activation, based on a window of width $3$ and height equal to the number of classes, and with a stride of one token. We use a wide convolution, by padding the input matrix to either size with vectors of zeros.
These $j$ feature maps are then subsampled with mean pooling, such that the network is easily able to capture the average uncertainty in each window. The final hidden layer $\mathbf{h}_e$ is used to represent the predictive marginals.

\paragraph{Confidence of sequential prediction}
The last component is a score $C$ which indicates the confidence of the model prediction. 
This is defined based on the most probable label sequence under the model, e.g., using Viterbi algorithm with a CRF, and the probability of this sequence is used to represent the confidence, $C = \sqrt[n]{\max_\mathbf{y} p_\phi(\mathbf{y}|\mathbf{x}_i)}$, where $n=|\mathbf{x}_i|$ is the length of the sentence.

\subsubsection{Action}
We now turn to the action, which denotes whether the human oracle must annotate the current sentence.
The agent selects either to annotate $\mathbf{x}_i$, in which case $a_i=1$, or not, with $a_i=0$, after which the agent proceeds to consider the next instance, $\mathbf{x}_{i+1}$.
When action $a_i=1$ is chosen, an oracle is requested to annotate the sentence, and the newly annotated sentence is added to the training data, and $\phi$ updated accordingly.
A special `terminate' option applies when no further data remains or the annotation budget is exhausted, which concludes the active learning run (referred to as an `episode' or `game' herein).
 

\subsubsection{Reward}
The training signal for learning the policy takes the form of a scalar `reward', which provides feedback on the quality of the  actions made by the agent.
The most obvious reward is to wait for a game to conclude, then measure the held-out performance of the model, which has been trained on the labelled data.
%
However, this reward is delayed, and is difficult to related to individual actions after a long game.
To compensate for this, we use reward shaping, whereby small intermediate rewards are assigned which speeds up the learning process~\cite{Ng2003shaping,lample2016playing}.
At each step, the intermediate reward is defined as the change in held-out performance, i.e., $R(s_{i-1},a)= \text{Acc}(\phi_i) - \text{Acc}(\phi_{i-1})$, where Acc denotes predictive accuracy (here F1 score), and $\phi_i$ is the trained model after action $a$ has take place, which may include an additional training instance.
Accordingly, when considering the aggregate reward over a game, the intermediate terms cancel, such that the total reward measures the performance improvement over the whole game.
Note that the value of $R(s,a)$ can be positive or negative,  indicating a beneficial or detrimental effect on the performance.

\subsubsection{Budget}
There is a fixed budget $\mathcal{B}$ for the total number of instances annotated, which corresponds to the terminal state in the MDP. It is a predefined number and chosen according to time and cost constraints. A game is finished when the data is exhausted or the budget reached, and with the final result being the dataset thus created, upon which the final model is trained.

\begin{algorithm}[t]
	\caption{Learn an active learning policy}
	\begin{algorithmic}[1]
		\REQUIRE data $\mathcal{D}$, budget $\mathcal{B}$
		\ENSURE  $\pi$
		
		\FOR{episode = $1,2,\dots, N$}
		
		\STATE $\mathcal{D}_l \leftarrow \emptyset$ and shuffle $\mathcal{D}$
		
		\STATE $\phi \leftarrow \text{Random}$ 
		
		\FOR{$i \in \{0,1,2,\dots,|\mathcal{D}|\}$}
		
		\STATE Construct the state $s_i$ using $\mathbf{x}_i$
		
		\STATE The agent makes a decision according to $a_i=\argmax{Q^\pi(s_i,a)}$
		
		\IF{$a_i =1$}
		\STATE Obtain the annotation $\mathbf{y}_i$
		\STATE $\mathcal{D}_l \leftarrow \mathcal{D}_l + (\mathbf{x}_i,\mathbf{y}_i)$
		\STATE Update model $\phi$ based on $\mathcal{D}_l$
		\ENDIF
		
		\STATE Receive a reward $r_i$ using held-out set
		
		\IF{$|\mathcal{D}_l| = \mathcal{B}$}
		\STATE Store $(s_i,a_i,r_i,\text{Terminate})$ in $\mathcal{M}$
		\STATE Break
		\ENDIF
		
		\STATE Construct the new state $s_{i+1}$
		
		\STATE Store transition $(s_i,a_i,r_i,s_{i+1})$ in $\mathcal{M}$
		
		\STATE Sample random minibatch of transitions $\{ (s_j,a_j,r_j,s_{j+1}) \}$ from $\mathcal{M}$, and perform gradient descent step on $\mathcal{L}(\theta)$
		
		
		\STATE Update policy $\pi$ with $\theta$
		
		\ENDFOR
		
		\ENDFOR
		\RETURN the latest policy $\pi$
	\end{algorithmic}
	\label{alg-learn-policy}
\end{algorithm}

\subsubsection{Reinforcement learning}
The remaining question is how the above components can be used to \emph{learn} a good policy.
Different policies make different data selections, and thus result in models with different performance.
We adopt a reinforcement learning (RL) approach to learn a policy resulting a highly accurate model.

Having represented the problem as a MDP, episode as a sequence of transitions $(s_i,a,r,s_{i+1})$. 
One episode of active learning produces a finite sequence of states, actions and rewards. 
We use a deep $Q$-learning approach~\cite{mnih2015human}, which formalises the policy using function $Q^{\pi}(s,a) \rightarrow \mathcal{R}$ which determines the utility of taking $a$ from state $s$ according to a policy $\pi$. 
In $Q$-learning, the agent iteratively updates $Q(s,a)$ using rewards obtained from each episode, with updates based on the recursive Bellman equation for the optimal $Q$:
\begin{equation}
Q^{\pi}(s,a) = \mathbb{E}[R_i |s_i =s, a_i=a, \pi].
\end{equation}
Here, $R_i=\sum_{t=i}^{T}\gamma^{t-i} r_t$ is the discounted future reward and $\gamma \in [0,1]$ is a factor discounting the value of future rewards and the expectation is taken over all transitions involving state $s$ and action $a$.

Following Deep Q-learning~\cite{mnih2015human}, we make use of a deep neural network to compute the expected Q-value, in order to update the parameters.
We implement the Q-function using a 
single hidden layer neural network, taking as input the state representation $(\mathbf{h}_c, \mathbf{h}_e, C)$ (defined in \S\ref{sec-state}),  and outputting two scalar values corresponding to the values $Q(s, a)$ for $a \in \{0,1\}$. 
This network uses a rectified linear unit (ReLU) activation function in its hidden layer.

The parameters in the DQN are learnt using stochastic gradient descent, based on a regression objective to match the $Q$-values predicted by the DQN and the expected Q-values from the Bellman equation, $r_i + \gamma \max_a Q(s_{i+1}, a; \theta)$. 
Following~\cite{mnih2015human}, we use an experience replay memory $\mathcal{M}$ to store each transition $(s,a,r,s')$ as it is used in an episode,  after which we sample a mini-batch of transitions from the memory and then minimize the loss function:
\begin{equation}
\mathcal{L}(\theta) = \mathbb{E}_{s,a, r, s'} \left[\left(y_i(r, s') -Q(s,a;\theta)\right)^2\right],
\label{eq-loss}
\end{equation}
where $y_i(r, s') = r + \gamma \max_{a'}Q(s',a';\theta_{i-1})$ is the target $Q$-value, based on the current parameters $\theta_{i-1}$, and the expectation is over the mini-batch. 
Learning updates are made every training step, based on stochastic gradient descent to minimise Eq.~\ref{eq-loss} w.r.t.\@ parameters $\theta$.


The algorithm for learning is summarised in Algorithm~\ref{alg-learn-policy}. We train the policy by running multiple active learning episodes over the training data, where each episode is a simulated active learning run. For each episode, we shuffle the data, and hide the known labels, which are revealed as requested during the run.  A disjoint held-out set is used to compute the reward, i.e., model accuracy, which is fixed over the episodes. Between each episode the model is reset to its initialisation condition, with the main changes being the different (random) data ordering and the evolving policy function.

\subsection{Cross-lingual policy transfer}
\begin{algorithm}[t]
	\caption{Active learning by policy transfer}
	\begin{algorithmic}[1]
		\REQUIRE unlabelled data $\mathcal{D}$,  budget $\mathcal{B}$, policy $\pi$
		\ENSURE  $\mathcal{D}_l$
		
		\STATE $\mathcal{D}_l \leftarrow \emptyset$
		
		\STATE $\phi \leftarrow \text{Random}$
		
		\FOR{$|\mathcal{D}_l| \neq \mathcal{B}$ and $\mathcal{D}$ not empty} 
		
		\STATE Randomly sample $\mathbf{x}_i$ from the data pool $\mathcal{D}$ and construct the state $s_i$
		
		\STATE The agent chooses an action $a_i$ according to $a_i=\argmax{Q^\pi(s_i,a)}$
		
		\IF{$a_i=1$}
		
		\STATE Obtain the annotation $\mathbf{y}_i$
		
		\STATE $\mathcal{D}_l \leftarrow \mathcal{D}_l + (\mathbf{x}_i,\mathbf{y}_i)$
		
		\STATE Update model $\phi$ based on $\mathcal{D}_l$
		
		\ENDIF

		\STATE $\mathcal{D} \leftarrow \mathcal{D} \backslash \mathbf{x}_i$

		\STATE Receive a reward $r_i$ using held-out set
		
		\STATE Update policy $\pi$
		
		\ENDFOR
		
		\RETURN $\mathcal{D}_l$
	\end{algorithmic}
	\label{alg-apply-policy}
\end{algorithm}

We now turn to the question of how the learned policy can be applied to another dataset.
Given the extensive use of the training dataset, the policy application only makes sense when employed in a different data setting, e.g., where the domain, task or language is different. 
For this paper, we consider a cross-lingual application of the same task (NER), where we train a policy on a source language (e.g., English), and then transfer the learned policy to a different target language.
Cross-lingual word embeddings provide a common shared representation to facilitate application of the policy to other languages.

We illustrate the policy transfer algorithm in Algorithm~\ref{alg-apply-policy}.
This algorithm is broadly similar to Algorithm~\ref{alg-learn-policy}, but has two key differences.
Firstly, Algorithm~\ref{alg-apply-policy} makes only one pass over the data, rather than several passes, as befits an application to a low-resource language where oracle labelling is costly.
Secondly, the algorithm also assumes an initial policy, $\pi$, which is fine tuned during the episode based on held-out performance such that  the policy can adapt to the test scenario.%
\footnote{Moreover, the algorithm can be extended to a traditional batch setting by evaluating a batch of data instances and selectinag the best $k$ instances for labelling under the policy. This could be applied in either the transfer step (Algorithm~\ref{alg-apply-policy}) or initial policy training (Algorithm~\ref{alg-learn-policy}), or both.}

\subsection{Cold-start transfer }

\begin{algorithm}[t]
	\caption{Active learning by policy \emph{and} model transfer, for `cold-start' scenario}
	\begin{algorithmic}[1]
		\REQUIRE unlabelled data $\mathcal{D}$, budget $\mathcal{B}$, policy $\pi$, model $\phi$
		\ENSURE  $\mathcal{D}_l$
		
		\STATE $\mathcal{D}_l \leftarrow \emptyset$
		\FOR{$|\mathcal{D}_l| \neq \mathcal{B}$ and $\mathcal{D}$ not empty}
		
		\STATE Randomly sample $\mathbf{x}_i$ from the data pool $\mathcal{D}$ and construct the state $s_i$
		
		\STATE The agent chooses an action $a_i$ according to $a_i=\argmax{Q^\pi(s_i,a)}$
		
		\IF{$a_i=1$}
		
		\STATE $\mathcal{D}_l \leftarrow \mathcal{D}_l + (\mathbf{x}_i,-)$
		
		\ENDIF

		\STATE $\mathcal{D} \leftarrow \mathcal{D} \backslash \mathbf{x}_i$

		\ENDFOR
		
		\STATE Obtain all the annotations for $\mathcal{D}_l$
		
		\RETURN $\mathcal{D}_l$
	\end{algorithmic}
	\label{alg-cold-al}
\end{algorithm}

The above transfer algorithm has some limitations, which may not be realistic for low-resource settings:
the requirement for held-out evaluation data and the embedding of the oracle annotator inside the learning loop. 
The former implies more supervision than is ideal in a low-resource setting, while the latter places limitations on the communication with annotator as well as a necessity for real-time processing, both which are unlikely in a field linguistics setting.

For this data and- communication-impoverished setting, denoted as \emph{cold-start}, we allow only one chance to request labels for the target data, and, having no held-out data, do not allow policy updates.
The agent needs to select a batch of unlabelled target instances for annotations, but cannot use these resulting annotations or any other feedback to refine the selection.
In this, more difficult cold-start setting, we bootstrap the process with an initial model, such that the agent can make informative decisions in the absence of feedback.

The procedure is outlined in Algorithm~\ref{alg-cold-al}.
Using the cross-lingual word embeddings, we transfer both a policy and a model into the target language.
The model, $\phi$, is trained on one source language, and the policy is learned on a different source language.
Policy learning uses Alg~\ref{alg-learn-policy}, with the small change that in step 3 the model is initialised using $\phi$.
Consequently the learned policy can exploit the knowledge from cross-lingual initialisation, such that it can figure out which aspects that need to be corrected using target annotated data.
Overall this allows for estimates and confidence values to be produced by the model,  thus providing the agent with sufficient information for data selection.



\section{Experiments}

We conduct experiments to validate the proposed active learning method in a cross-lingual setting, whereby an active learning policy trained on a source language is transferred to a target language.
We allow repeated active learning simulations on the source language, where annotated corpora are plentiful, to learn a policy, while for target languages we only permit a single episode, to mimic a language without existing resources.


\begin{figure*}
	\includegraphics[width=1.05\linewidth]{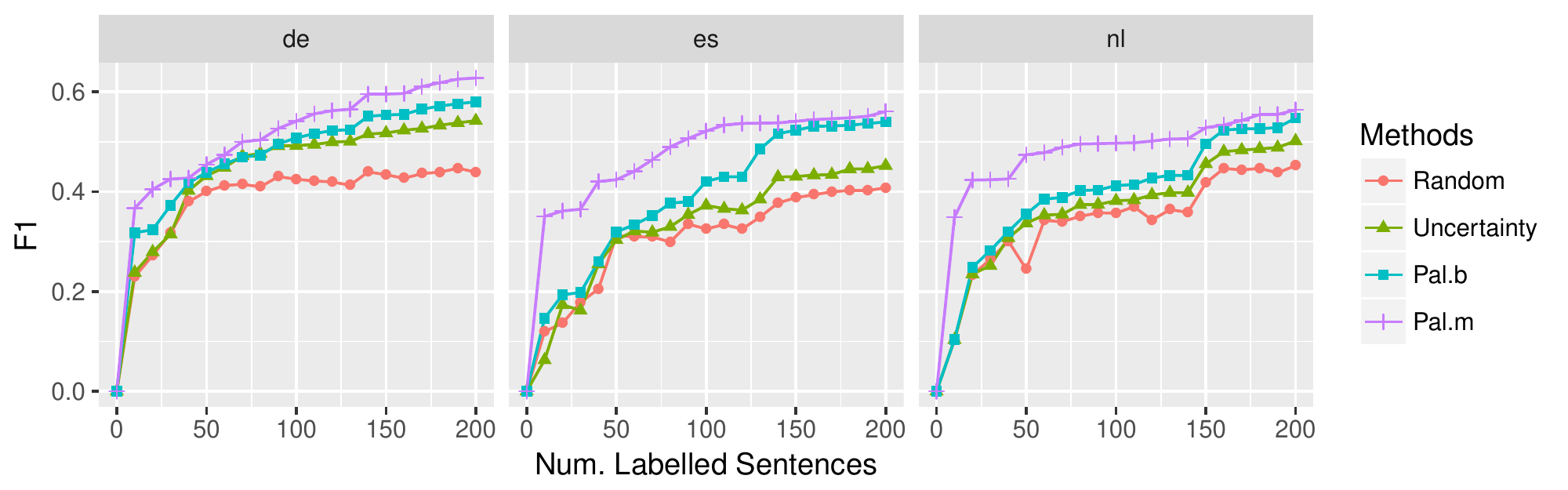}
	
	
	\caption{The performance of active learning methods on the bilingual and multilingual settings for three target languages, whereby the active learning policy is trained on only en, or all other languages excluding the target, respectively.}
	\label{fig-standard}
\end{figure*}

We use NER corpora from CoNLL2002/2003 shared tasks,\footnote{\label{conll2002} \url{http://www.cnts.ua.ac.be/conll2002/ner/}, \url{http://www.cnts.ua.ac.be/conll2003/ner/}}
which comprise NER annotated text in English (en), German (de), Spanish (es), and Dutch (nl), each annotated using the IOB1 labelling scheme, which we convert to the IO labeling scheme. 
We use the existing corpus partions, with \texttt{train} used for policy training, \texttt{testb} used as held-out for computing rewards, and final results are reported on \texttt{testa}.

We consider three experimental conditions, as illustrated in Table~\ref{tab-data}:
\begin{description}
\item[bilingual] where English is the source (used for policy learning) and we vary the target language;
\item[multilingual] where several source languages are the used in joint learning of the policy, and a separate language is used as target; and
\item[cold-start] where a pretrained English NER tagger is used to initialise policy learning on a source language, and in cold-start application to a separate target language.
\end{description} 


\begin{table}
	\begin{center}
		\begin{tabular}{ c c|c  c|c c  c}
			\hline
			\multicolumn{2}{c|}{Bilingual} & \multicolumn{2}{c|}{Multilingual} & \multicolumn{3}{c}{Cold-start} \\
			\hline
			\textbf{tgt} & \textbf{src} & \textbf{tgt} & \textbf{src} & \textbf{tgt} & \textbf{src} & \textbf{pre} \\  \hline
			de & en & de & en,nl,es & de & nl & en \\
			nl & en & nl & en,de,es & nl & de & en \\
			es & en & es & en,de,nl & es & de & en \\
			- & - & - &  - & de & es & en \\
			- & - & - & - & nl & es & en \\
			- & - & - & - & es & nl & en \\
			\hline
			
		\end{tabular}
		\caption{Experimental configuration for the three settings, showing target language (\textbf{tgt}), source language (\textbf{src}) as used for policy learning, and language used for pre-training the model (\textbf{pre}).}
		\label{tab-data}
	\end{center}
\end{table}

\paragraph{Configuration}

We now outline the parameter settings for the experimental runs.
For learning an active learning policy, we run $N=10,000$ episodes with budget $\mathcal{B}=200$ sentences using  Alg.~\ref{alg-learn-policy}.
Content representations use three convolutional filters of size $3,4$ and $5$, using $128$ filters for each size, while for predictive marginals, the convolutional filters are of width $3$, using $20$ filters.
The size of the last hidden layer is $256$. 
The discount factor is set to $\gamma = 0.99$.
We used the ADAM algorithm with mini-batches of size $32$ for training the neural network.
To report performance, we apply the learned policy to the target training set (using Alg.~\ref{alg-apply-policy} or~\ref{alg-cold-al}, again with budget 200),\footnote{Although it is possible the policy may learn not to use the full budget, this does not occur in practise.}  after which we use the final trained model for which we report $F_1$ score.

For word embeddings, we use off the shelf CCA trained multilingual embeddings \cite{ammar2016massively},\footnote{\url{http://128.2.220.95/multilingual}} using a $40$ dimensional embedding and fixing these during training of both the policy and model.
As the model, we use a standard linear chain CRF \cite{lafferty2001crf} for the first two sets of experiments, while for cold-start case we use a basic RNN classifier with the same multilingual embeddings as before, and  a 128 dimensional hidden layer.

The proposed method is referred to as \pal, as shorthand \emph{Policy based Active Learning}.
Subscripts $b,m,c$ are used to denote the bilingual, multilingual and cold-start experimental configurations. 
For comparative baselines, we use the following methods: 
\begin{description}
	\item[Uncertainty sampling] we use the total token entropy measure~\cite{settles2008analysis}, which takes the instance $\mathbf{x}$ maximising $\sum_{t=1}^{|\mathbf{x}|} H(y_t | \mathbf{x}, \phi),$ where $H$ is the token entropy.  We use the whole training set as the data pool, and select a single instance for labelling in each active learning step.
        This method was shown to achieve the best result among model-independent active learning methods on the CoNLL data. 
	\item[Random sampling] which randomly selects examples from the unlabelled pool.
\end{description}

\begin{figure*}
	\includegraphics[width=1.05\linewidth]{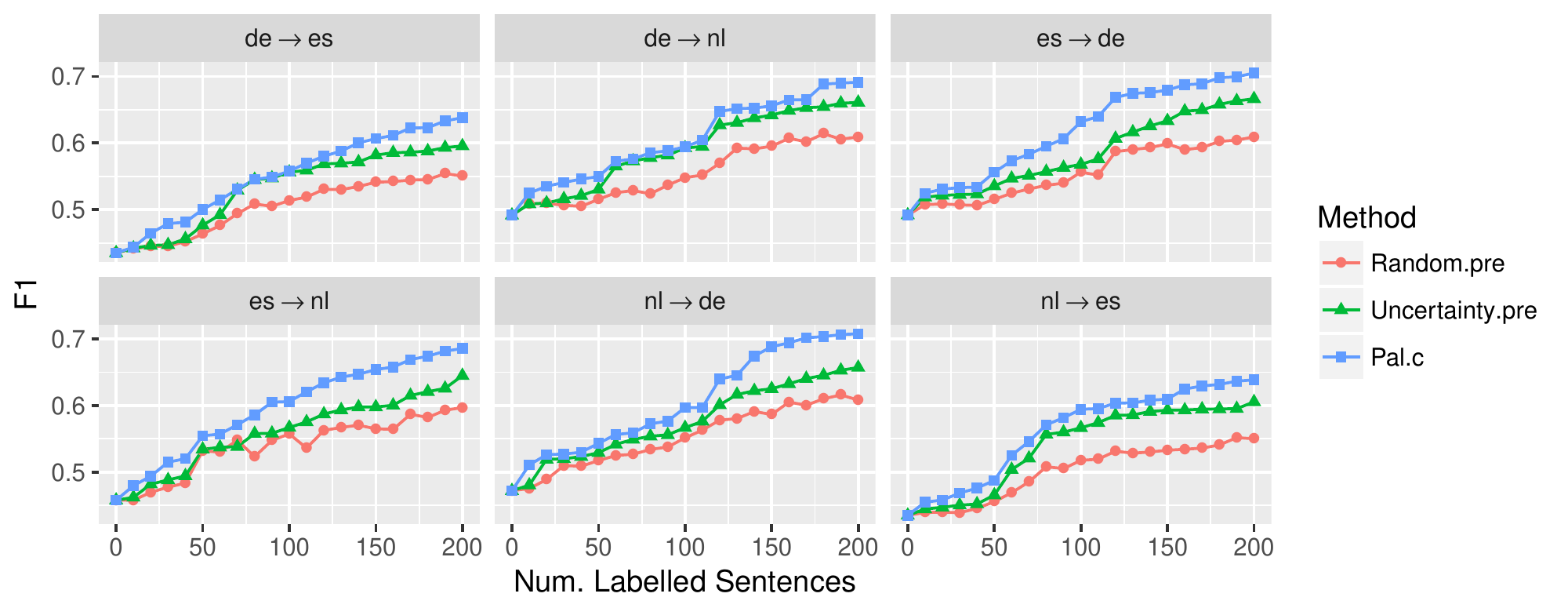}
	\caption{The performance of active learning methods on the cold-start setting, each showing different source $\rightarrow$ target configurations, in all cases pretraining in en.}
	\label{fig-cold}
\end{figure*}

\paragraph{Results}
Figure~\ref{fig-standard} shows results the bilingual case, where \palb consistently outperforms the Random and Uncertainty baselines across the three target languages.  Uncertainty sampling is ineffective, particularly towards the start of the run, as a consequence of its dependence on a high quality model. The use of content information allows \palb to make a stronger start, despite the poor initial model.

Also shown in Figure~\ref{fig-standard} are results for multilingual policy learning, \palm, which outperform all other approaches including \palb. This illustrates that the additional training over several languages gives rise to a better policy, than only using one source language. The superior performance is particularly marked in the early stages of the runs for Spanish and Dutch, which may indicate that the approach was better able to learn to exploit the sentence content information.  

We evaluate the cold-start setting in Figure~\ref{fig-cold}.
Recall that in this setting there are no  policy or model updates, as no heldout data is used, and all annotations arrive in a batch.
The model, however, is initialised with a NER tagger trained on a different language, which explains why the performance for all methods starts from around $40\%$ rather than $0\%$.
Even in this challenging evaluation setting, our algorithm \palc outperforms both baseline methods, showing that deep Q learning allows for better exploitation of the pretrained classifier, alongside the sentence content.

Lastly, we report the results for all approaches in Table~\ref{tab-200}, based on training on the full $200$ labelled sentences as selected under the different methods. It is clear that the \pal methods all outperform the baselines, and among these the multilingual training of \palm outperforms the bilingual setting in \palb. Surprisingly, \palc gives the overall best results, despite using a static policy and model during target application, underscoring the importance of model pretraining. 
Table~\ref{tab-200} also reports the cost reduction versus random sampling, showing that the \pal methods can reduce the annotation burden to as low as $10\%$.

\begin{table}
	\sisetup{round-mode=places,round-precision=1,table-format=2.1}
	\begin{center}
		\begin{tabular}{  c S S[round-precision=0,table-format=2.0]  S S[round-precision=0,table-format=2.0]  S S[round-precision=0,table-format=2.0]}
			\toprule
			~ & \multicolumn{2}{c}{\bf de} & \multicolumn{2}{c}{\bf nl} & \multicolumn{2}{c}{\bf  es} \\
			~  & {F1} & {C/R} &  {F1} & {C/R}  & {F1} & {C/R}  \\
			\midrule
			Rand. & 44.6 & 100 &  45.2 & 100 & 40.7 & 100  \\
			Uncert. & 54.2 & 60 & 50.1 & 25 & 45.1 &  30 \\
			\palb & 57.9 & 60 & 54.7 & 25 & 53.9  & 40 \\
			\palm  & 62.7 & 25 & 56.3 & 30 & 56.0 & 25 \\
			\palc & 70.7 & 10 & 69.1 & 10 & 63.8 & 10 \\
			\bottomrule
		\end{tabular}
		\caption{Results from active learning using the different methods, where each approach constructs a training set of 200 sentences. The three target languages are shown as columns, reporting in each F$_1$ score (\%) and the relative cost reduction to match the stated performance of the Random strategy.}
		\label{tab-200}
	\end{center}
\end{table}

\section{Conclusion}
In this paper, we have proposed a new active learning algorithm capable of learning active learning strategies from data. We formalise active learning under a Markov decision framework, whereby active learning corresponds to a sequence of binary annotation decisions applied to a stream of data. Based on this, we design an active learning algorithm as a policy based on deep reinforcement learning.
We show how these learned active learning policies can be transferred between languages,
which we empirically show provides consistent and sizeable improvements over baseline methods, including traditional uncertainty sampling.
This holds true  even in a very difficult cold-start setting, where no evaluation data is available, and there is no ability to react to annotations.

\section*{Acknowledgments}
This work was sponsored by the Defense Advanced Research Projects Agency Information Innovation Office (I2O) under the Low Resource Languages for Emergent Incidents (LORELEI) program issued by DARPA/I2O under Contract No. HR0011-15-C-0114. The views expressed are those of the authors and do not reflect the official policy or position of the Department of Defense or the U.S. Government. Trevor Cohn was supported by an Australian Research Council Future Fellowship.

\bibliography{emnlp2017}

\begin{thebibliography}{}
\expandafter\ifx\csname natexlab\endcsname\relax\def\natexlab#1{#1}\fi

\bibitem[{Ammar et~al.(2016)Ammar, Mulcaire, Tsvetkov, Lample, Dyer, and
  Smith}]{ammar2016massively}
Waleed Ammar, George Mulcaire, Yulia Tsvetkov, Guillaume Lample, Chris Dyer,
  and Noah~A Smith. 2016.
\newblock Massively multilingual word embeddings.
\newblock {\em arXiv preprint arXiv:1602.01925\/} .

\bibitem[{Ba et~al.(2015)Ba, Mnih, and Kavukcuoglu}]{ba2014multiple}
Jimmy Ba, Volodymyr Mnih, and Koray Kavukcuoglu. 2015.
\newblock Multiple object recognition with visual attention.
\newblock In {\em Proceedings of the International Conference on Learning
  Representations (ICLR)\/}.

\bibitem[{Bengio(2012)}]{bengio2012deep}
Yoshua Bengio. 2012.
\newblock Deep learning of representations for unsupervised and transfer
  learning.
\newblock In {\em Proceedings of ICML Workshop on Unsupervised and Transfer
  Learning\/}. pages 17--36.

\bibitem[{Fang and Cohn(2017)}]{fang2017model}
Meng Fang and Trevor Cohn. 2017.
\newblock Model transfer for tagging low-resource languages using a bilingual
  dictionary.
\newblock In {\em Proceedings of the 55th Annual Meeting on Association for
  Computational Linguistics (ACL)\/}.

\bibitem[{Garrette and Baldridge(2013)}]{garrette2013learning}
Dan Garrette and Jason Baldridge. 2013.
\newblock Learning a part-of-speech tagger from two hours of annotation.
\newblock In {\em Proceedings of the 2013 Conference of the North American
  Chapter of the Association for Computational Linguistics: Human Language
  Technologies (HLT-NAACL)\/}. pages 138--147.

\bibitem[{Kim(2014)}]{kim2014convolutional}
Yoon Kim. 2014.
\newblock Convolutional neural networks for sentence classification.
\newblock In {\em Proceedings of the 2014 Conference on Empirical Methods on
  Natural Language Processing (EMNLP)\/}.

\bibitem[{Lafferty et~al.(2001)Lafferty, McCallum, and
  Pereira}]{lafferty2001crf}
John~D. Lafferty, Andrew McCallum, and Fernando C.~N. Pereira. 2001.
\newblock Conditional random fields: Probabilistic models for segmenting and
  labeling sequence data.
\newblock In {\em Proceedings of the Eighteenth International Conference on
  Machine Learning (ICML)\/}. pages 282--289.

\bibitem[{Lample and Chaplot(2016)}]{lample2016playing}
Guillaume Lample and Devendra~Singh Chaplot. 2016.
\newblock Playing {FPS} games with deep reinforcement learning.
\newblock {\em arXiv preprint arXiv:1609.05521\/} .

\bibitem[{Levine et~al.(2016)Levine, Finn, Darrell, and Abbeel}]{levine2016end}
Sergey Levine, Chelsea Finn, Trevor Darrell, and Pieter Abbeel. 2016.
\newblock End-to-end training of deep visuomotor policies.
\newblock {\em Journal of Machine Learning Research\/} 17(39):1--40.

\bibitem[{Lewis and Gale(1994)}]{lewis1994sequential}
David~D Lewis and William~A Gale. 1994.
\newblock A sequential algorithm for training text classifiers.
\newblock In {\em Proceedings of the 17th International ACM SIGIR Conference on
  Research and Development in Information Retrieval\/}. pages 3--12.

\bibitem[{Maddison et~al.(2015)Maddison, Huang, Sutskever, and
  Silver}]{maddison2014move}
Chris~J Maddison, Aja Huang, Ilya Sutskever, and David Silver. 2015.
\newblock Move evaluation in go using deep convolutional neural networks.
\newblock In {\em Proceedings of the International Conference on Learning
  Representations (ICLR)\/}.

\bibitem[{McCallumzy and Nigamy(1998)}]{mccallumzy1998employing}
Andrew~Kachites McCallumzy and Kamal Nigamy. 1998.
\newblock Employing em and pool-based active learning for text classification.
\newblock In {\em Proceedings of the 15th International Conference on Machine
  Learning (ICML)\/}. pages 359--367.

\bibitem[{Mnih et~al.(2015)Mnih, Kavukcuoglu, Silver, Rusu, Veness, Bellemare,
  Graves, Riedmiller, Fidjeland, Ostrovski et~al.}]{mnih2015human}
Volodymyr Mnih, Koray Kavukcuoglu, David Silver, Andrei~A Rusu, Joel Veness,
  Marc~G Bellemare, Alex Graves, Martin Riedmiller, Andreas~K Fidjeland, Georg
  Ostrovski, et~al. 2015.
\newblock Human-level control through deep reinforcement learning.
\newblock {\em Nature\/} 518(7540):529--533.

\bibitem[{Nair et~al.(2015)Nair, Srinivasan, Blackwell, Alcicek, Fearon,
  De~Maria, Panneershelvam, Suleyman, Beattie, Petersen, Legg, Mnih,
  Kavukcuoglu, and Silver}]{nair2015massively}
Arun Nair, Praveen Srinivasan, Sam Blackwell, Cagdas Alcicek, Rory Fearon,
  Alessandro De~Maria, Vedavyas Panneershelvam, Mustafa Suleyman, Charles
  Beattie, Stig Petersen, Shane Legg, Volodymyr Mnih, Koray Kavukcuoglu, and
  David Silver. 2015.
\newblock Massively parallel methods for deep reinforcement learning.
\newblock In {\em Proceedings of ICML Workshop on Deep Learning\/}.

\bibitem[{Narasimhan et~al.(2016)Narasimhan, Yala, and
  Barzilay}]{narasimhan2016improving}
Karthik Narasimhan, Adam Yala, and Regina Barzilay. 2016.
\newblock Improving information extraction by acquiring external evidence with
  reinforcement learning.
\newblock In {\em Proceedings of the 2016 Conference on Empirical Methods on
  Natural Language Processing (EMNLP)\/}.

\bibitem[{Ng(2003)}]{Ng2003shaping}
Andrew~Y. Ng. 2003.
\newblock {\em Shaping and Policy Search in Reinforcement Learning\/}.
\newblock Ph.D. thesis, University of California, Berkeley.

\bibitem[{Parisotto et~al.(2016)Parisotto, Ba, and
  Salakhutdinov}]{parisotto2015actor}
Emilio Parisotto, Jimmy~Lei Ba, and Ruslan Salakhutdinov. 2016.
\newblock Actor-mimic: Deep multitask and transfer reinforcement learning.
\newblock In {\em Proceedings of the International Conference on Learning
  Representations (ICLR)\/}.

\bibitem[{Qian et~al.(2014)Qian, Hui, Hu, Zhou, and Zhu}]{Qian2014BilingualAL}
Longhua Qian, Haotian Hui, Yanan Hu, Guodong Zhou, and Qiaoming Zhu. 2014.
\newblock Bilingual active learning for relation classification via pseudo
  parallel corpora.
\newblock In {\em Proceedings of the 52nd Annual Meeting of the Association for
  Computational Linguistics (ACL)\/}. pages 582--592.

\bibitem[{Roy and McCallum(2001)}]{roy2001toward}
Nicholas Roy and Andrew McCallum. 2001.
\newblock Toward optimal active learning through monte carlo estimation of
  error reduction.
\newblock In {\em Proceedings of the 18th International Conference on Machine
  Learning (ICML)\/}. pages 441--448.

\bibitem[{Rusu et~al.(2016)Rusu, Rabinowitz, Desjardins, Soyer, Kirkpatrick,
  Kavukcuoglu, Pascanu, and Hadsell}]{rusu2016progressive}
Andrei~A Rusu, Neil~C Rabinowitz, Guillaume Desjardins, Hubert Soyer, James
  Kirkpatrick, Koray Kavukcuoglu, Razvan Pascanu, and Raia Hadsell. 2016.
\newblock Progressive neural networks.
\newblock {\em arXiv preprint arXiv:1606.04671\/} .

\bibitem[{Settles(2010)}]{settles2010active}
Burr Settles. 2010.
\newblock Active learning literature survey.
\newblock {\em University of Wisconsin, Madison\/} 52(55-66):11.

\bibitem[{Settles and Craven(2008)}]{settles2008analysis}
Burr Settles and Mark Craven. 2008.
\newblock An analysis of active learning strategies for sequence labeling
  tasks.
\newblock In {\em Proceedings of the Conference on Empirical Methods in Natural
  Language Processing (EMNLP)\/}. pages 1070--1079.

\bibitem[{Seung et~al.(1992)Seung, Opper, and Sompolinsky}]{seung1992query}
H~Sebastian Seung, Manfred Opper, and Haim Sompolinsky. 1992.
\newblock Query by committee.
\newblock In {\em Proceedings of the 5th annual workshop on Computational
  Learning Theory\/}. pages 287--294.

\bibitem[{Shen et~al.(2004)Shen, Zhang, Su, Zhou, and Tan}]{shen2004multi}
Dan Shen, Jie Zhang, Jian Su, Guodong Zhou, and Chew-Lim Tan. 2004.
\newblock Multi-criteria-based active learning for named entity recognition.
\newblock In {\em Proceedings of the 42nd Annual Meeting on Association for
  Computational Linguistics (ACL)\/}.

\bibitem[{Silver et~al.(2016)Silver, Huang, Maddison, Guez, Sifre, Van
  Den~Driessche, Schrittwieser, Antonoglou, Panneershelvam, Lanctot
  et~al.}]{silver2016mastering}
David Silver, Aja Huang, Chris~J Maddison, Arthur Guez, Laurent Sifre, George
  Van Den~Driessche, Julian Schrittwieser, Ioannis Antonoglou, Veda
  Panneershelvam, Marc Lanctot, et~al. 2016.
\newblock Mastering the game of go with deep neural networks and tree search.
\newblock {\em Nature\/} 529(7587):484--489.

\bibitem[{Stratos and Collins(2015)}]{stratos2015simple}
Karl Stratos and Michael Collins. 2015.
\newblock Simple semi-supervised pos tagging.
\newblock In {\em Proceedings of the 2015 Conference of the North American
  Chapter of the Association for Computational Linguistics: Human Language
  Technologies (NAACL-HLT)\/}. pages 79--87.

\bibitem[{Thompson et~al.(1999)Thompson, Califf, and
  Mooney}]{thompson1999active}
Cynthia~A Thompson, Mary~Elaine Califf, and Raymond~J Mooney. 1999.
\newblock Active learning for natural language parsing and information
  extraction.
\newblock In {\em Proceedings of the 16th International Conference on Machine
  Learning (ICML)\/}. pages 406--414.

\bibitem[{Tomanek et~al.(2007)Tomanek, Wermter, and Hahn}]{tomanek2007approach}
Katrin Tomanek, Joachim Wermter, and Udo Hahn. 2007.
\newblock An approach to text corpus construction which cuts annotation costs
  and maintains reusability of annotated data.
\newblock In {\em Proceedings of the 2007 Joint Conference on Empirical Methods
  in Natural Language Processing and Computational Natural Language Learning
  (EMNLP-CoNLL)\/}. pages 486--495.

\bibitem[{Tong and Koller(2001)}]{tong2001support}
Simon Tong and Daphne Koller. 2001.
\newblock Support vector machine active learning with applications to text
  classification.
\newblock {\em Journal of Machine Learning Research\/} 2(Nov):45--66.

\bibitem[{Wang et~al.(2016)Wang, Schaul, Hessel, van Hasselt, Lanctot, and
  de~Freitas}]{wang2015dueling}
Ziyu Wang, Tom Schaul, Matteo Hessel, Hado van Hasselt, Marc Lanctot, and Nando
  de~Freitas. 2016.
\newblock Dueling network architectures for deep reinforcement learning.
\newblock In {\em Proceedings of the 33rd International Conference on Machine
  Learning (ICML)\/}.

\bibitem[{Watter et~al.(2015)Watter, Springenberg, Boedecker, and
  Riedmiller}]{watter2015embed}
Manuel Watter, Jost Springenberg, Joschka Boedecker, and Martin Riedmiller.
  2015.
\newblock Embed to control: A locally linear latent dynamics model for control
  from raw images.
\newblock In {\em Advances in Neural Information Processing Systems (NIPS)\/}.
  pages 2746--2754.

\bibitem[{Zhang et~al.(2016)Zhang, Pan, Wang, Vaswani, Ji, Knight, and
  Marcu}]{zhang2016name}
Boliang Zhang, Xiaoman Pan, Tianlu Wang, Ashish Vaswani, Heng Ji, Kevin Knight,
  and Daniel Marcu. 2016.
\newblock Name tagging for low-resource incident languages based on
  expectation-driven learning.
\newblock In {\em Proceedings of the 2016 Conference of the North American
  Chapter of the Association for Computational Language Technologies
  (NAACL-HLT)\/}. pages 249--259.

\end{thebibliography}
\bibliographystyle{emnlp_natbib}

\end{document}